\documentclass[a4paper,conference]{IEEEtran}
\usepackage{graphicx}
\usepackage{amsmath}
\usepackage{amssymb}
\usepackage{pifont}
\newcommand{\cmark}{\ding{51}}%
\newcommand{\xmark}{\ding{55}}%
\usepackage{booktabs}
\usepackage{multirow}
\usepackage{graphicx}
\usepackage[inline]{enumitem}
\usepackage{booktabs}
\usepackage{hyperref}
\usepackage{xcolor}
\usepackage{caption}
\usepackage{subcaption}

\usepackage{soul}
\usepackage{cite}

\ifCLASSINFOpdf
\else
\fi
\usepackage{amsmath}
\usepackage{amssymb}
\usepackage{url}

\usepackage[utf8]{inputenc}
\usepackage[capitalize]{cleveref}
\crefname{section}{Sec.}{Secs.}
\Crefname{section}{Section}{Sections}
\Crefname{table}{Table}{Tables}
\crefname{table}{Tab.}{Tabs.}

\begin{document}
%
\title{MAP-Gen: An Automated 3D-Box Annotation Flow with Multimodal Attention Point Generator}

\author{\IEEEauthorblockN{Chang Liu, Xiaoyan Qian, Xiaojuan Qi, Edmund Y. Lam, Siew-Chong Tan, Ngai Wong}
\IEEEauthorblockA{The University of Hong Kong\\Pokfulam, Hong Kong\\
Email: lcon7@connect.hku.hk}}


%


\maketitle

\begin{abstract}
Manually annotating 3D point clouds is laborious and costly, limiting the training data preparation for deep learning in real-world object detection. While a few previous studies tried to automatically generate 3D bounding boxes from weak labels such as 2D boxes, the quality is sub-optimal compared to human annotators. This work proposes a novel autolabeler, called multimodal attention point generator (MAP-Gen), that generates high-quality 3D labels from weak 2D boxes. It leverages dense image information to tackle the sparsity issue of 3D point clouds, thus improving label quality. For each 2D pixel, MAP-Gen predicts its corresponding 3D coordinates by referencing context points based on their 2D semantic or geometric relationships. The generated 3D points densify the original sparse point clouds, followed by an encoder to regress 3D bounding boxes. Using MAP-Gen, object detection networks that are weakly supervised by 2D boxes can achieve 94$\sim$99\% performance of those fully supervised by 3D annotations. It is hopeful this newly proposed MAP-Gen autolabeling flow can shed new light on utilizing multimodal information for enriching sparse point clouds.
\end{abstract}


%
\IEEEpeerreviewmaketitle

\begin{figure}[ht]
    \centering
    \begin{subfigure}[b]{\columnwidth}
         \centering
         \includegraphics[width=\textwidth]{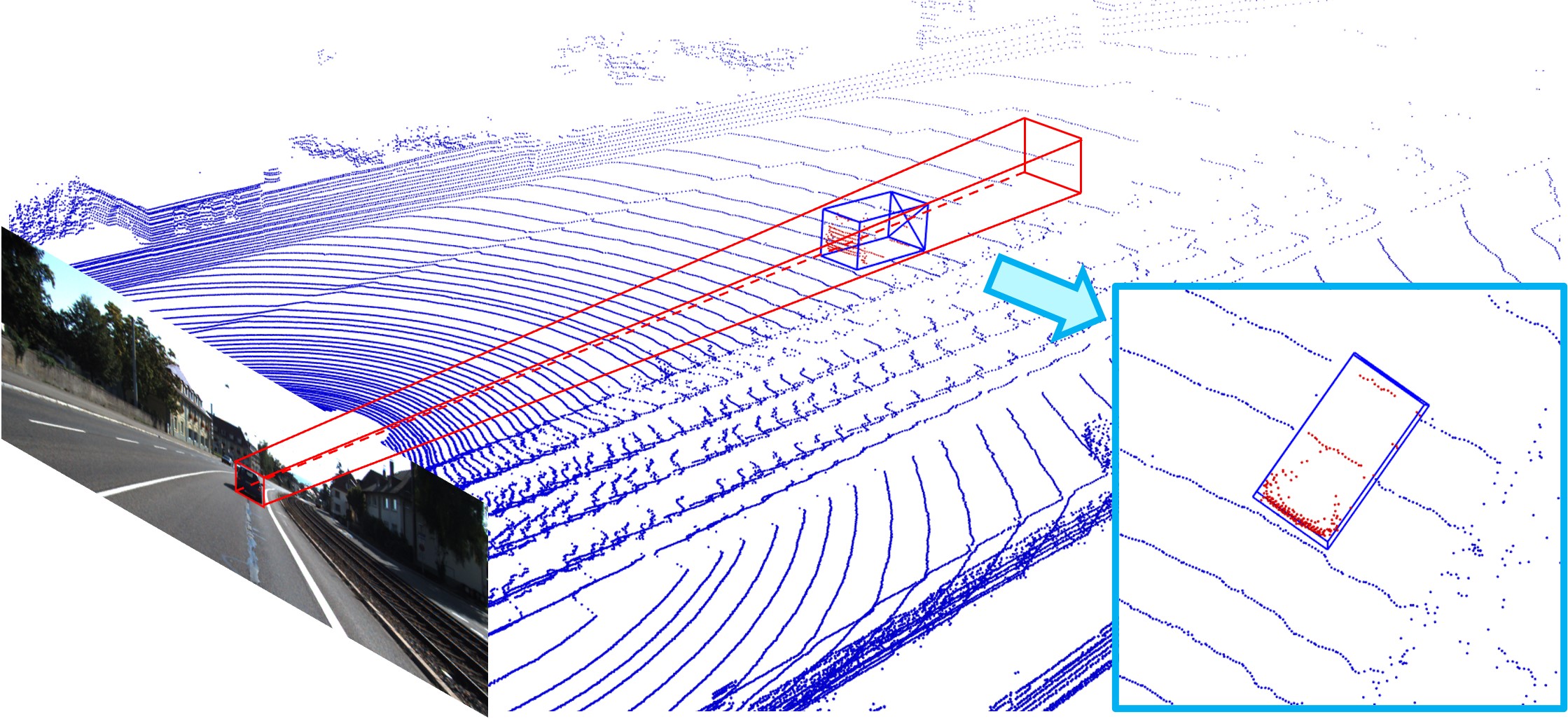}
         \caption{Frustum point cloud corresponding to a 2D box on image.}
         \label{fig:1a_frustum_projection}
     \end{subfigure}
     
    \begin{subfigure}[b]{0.49\columnwidth}
         \centering
         \includegraphics[width=\textwidth]{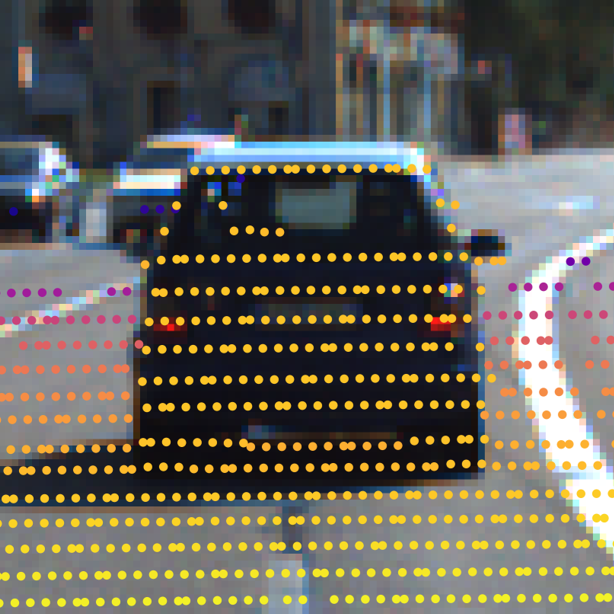}
         \caption{LiDAR-image projection.}
         \label{fig:1b_depth_img}
     \end{subfigure}
    \begin{subfigure}[b]{0.49\columnwidth}
         \centering
         \includegraphics[width=\textwidth]{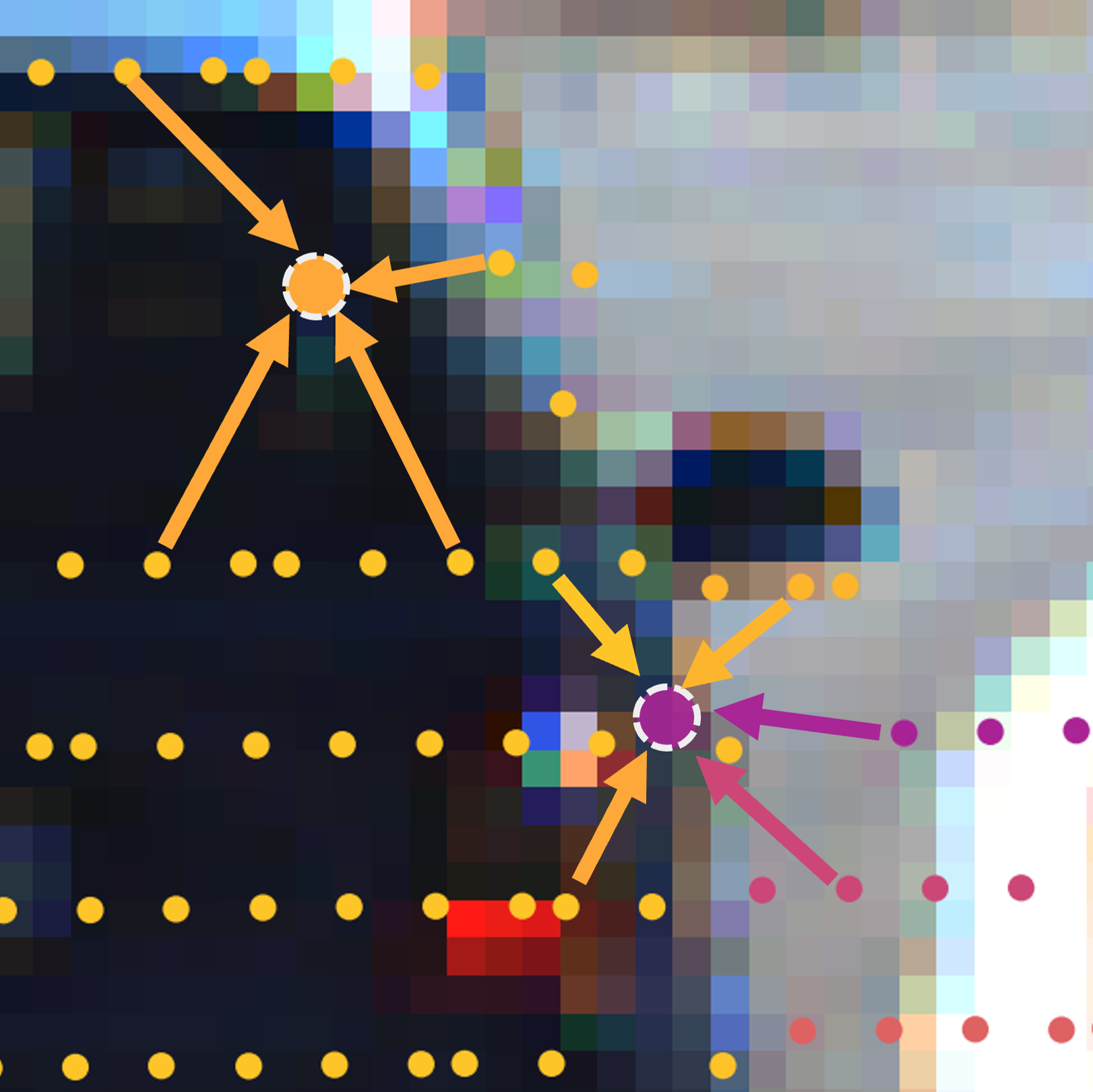}
         \caption{Image-guided interpolation.}
         \label{fig:1c_MAP-Gen}
     \end{subfigure}

    \caption{3D tasks are challenging due to the sparsity and irregularity of LiDAR scans. MAP-Gen resolves these by: (a) retrieving frustum point clouds from 2D boxes using calibration matrices; (b) projecting the LiDAR points onto images; (c) estimating a pixel's 3D coordinates from its context points based on 2D semantic and geometric relationships (represented in different colors and arrows). This last step is unique to MAP-Gen that sets its performance apart from existing approaches by a large margin.}
    \label{fig:fig1_intro}
\end{figure}

\section{Introduction}
Recently, 3D object detection models and datasets are emerging rapidly, especially in the context of autonomous driving. For open 3D detection datasets, such as KITTI Benchmark Dataset~\cite{Geiger2012CVPR_KITTI}, Waymo Open Dataset~\cite{sun2020scalability_waymo}, nuScenes Dataset~\cite{caesar2020nuscenes} and PandaSet~\cite{pandaset}, there is a trend for the datasets to grow larger to include various scenes. Also, fine-tuning on customized datasets is still a standard and effective practice in real-world applications. However, annotating these datasets requires enormous manual work, which limits deep learning techniques for broader application and deployment.

As is well-known, annotation requires much more human effort than data collection, particularly for LiDAR scans. To label a 3D bounding box, annotators need to pan and rotate the viewpoint repetitively, followed by adjusting the boxes. Statistics show it takes an experienced annotator around 114s for labeling a 3D bounding box, or 30s if an extra 3D object detector is used as an assistance~\cite{huang2019apolloscape, song2015sun_rgbd, meng2020weakly}. On the other hand, weak annotations such as 2D bounding boxes on images cost much less, which makes it promising if 3D boxes can be automatically generated from the weak annotations.

Several previous studies deploy weak annotations, such as center-clicks~\cite{meng2020weakly, bearman2016s_centerclick}, extreme-points~\cite{maninis2018deep_extremecut, papadopoulos2017extreme_click}, polygon-points~\cite{castrejon2017polygon_rnn, acuna2018polygon_rnnpp}, 2D bounding boxes~\cite{wei2021fgr, qi2018frustumpnet}, or 2D segmentation masks~\cite{mccraith2021lifting2d, wilson20203d_for_free_hdmap}. To facilitate 3D object detection with lower annotation costs, a few existing methods enhance the weak labels into a stronger form, for example, from 2D bounding boxes to 3D boxes~\cite{wei2021fgr}. After that, object detectors can be trained under weak supervision (i.e., 2D boxes), given 3D pseudo labels generated by these autolabelers.

However, the sparsity and irregularity of LiDAR scans make automatic 3D bounding box generation a challenging problem. As in~\cref{fig:1a_frustum_projection}, point clouds are sparse, especially for far regions where objects can be unrecognizable due to inadequate points. Since the sparsity grows as the distance gets farther, the number of points per instance can fluctuate significantly, thus complicating the 3D label generation task.

To handle such sparsity problem, some existing methods~\cite{qi2018frustumpnet, vora2020pointpainting, wei2021fgr} leverage RGB images to coarsely locate or distinguish objects in the sparse point clouds, as shown in~\cref{fig:1a_frustum_projection} and~\cref{fig:1b_depth_img}. For example, it is easier to distinguish objects from similarly shaped background objects (e.g., pedestrian v.s. traffic light pole) in images than in LiDAR point clouds~\cite{vora2020pointpainting}. However, almost all existing multimodal autolabelers~\cite{wei2021fgr, meng2020weakly, meng2021ws3d_towards, qin2020vs3d, mccraith2021lifting2d} take images as auxiliary, either cropping/zooming to particular regions of the point cloud or fusing point features with image features. Although they are the current state-of-the-art methods, we argue this approach does not fully address the sparsity problem because the cloud sizes remain unchanged. Their performance also exhibits an obvious gap versus human annotations.

As for the irregularity, voxel-based approaches~\cite{lang2019pointpillars, zhou2018voxelnet} convert point clouds into voxels for batch processing, but suffer from information loss. Alternatively, point-cloud-based methods~\cite{qi2017pointnet, qi2017pointnet++} randomly sample or duplicate points for uniform cloud sizes, which slightly affect the point distribution but without information loss. Nonetheless, the sampling strategy increases the computation overhead without substantial gain in accuracy as no extra information is introduced.

This paper proposes the multimodal attention point generator (MAP-Gen) that generates high-quality pseudo 3D annotations from 2D bounding boxes. We design a novel multimodal attention module to address the aforementioned issues simultaneously. For each 2D bounding box, the corresponding frustum region of the point cloud is extracted as in~\cref{fig:1a_frustum_projection}. Segmentation is then carried out on both the image and frustum point cloud to remove the backgrounds. After that, the multimodal attention module leverages image information to generate new foreground points and enrich the sparse point cloud. As shown in~\cref{fig:1c_MAP-Gen}, for each target pixel, MAP-Gen estimates the corresponding 3D coordinates by referring to other available context points, based on their semantic and geometric relationships on the image. In the last stage, we directly predict the 3D bounding box from the enriched point cloud. Experiments prove that MAP-Gen effectively improves the label quality, especially for difficult samples. Our contributions are threefold:
\begin{itemize}
    \item We develop MAP-Gen to automatically generate high-quality 3D annotations from weak 2D boxes, thus significantly saving human workloads.
    \item The proposed multimodal attention module constitutes a new way of leveraging image information to solve the sparsity and irregularity problems in LiDAR tasks.
    \item MAP-Gen is extensively evaluated on the KITTI Benchmark dataset, and shown to outperform existing state-of-the-art baselines by a large margin.
\end{itemize}

\begin{figure*}[t]
    \centering
    \includegraphics[width=\textwidth]{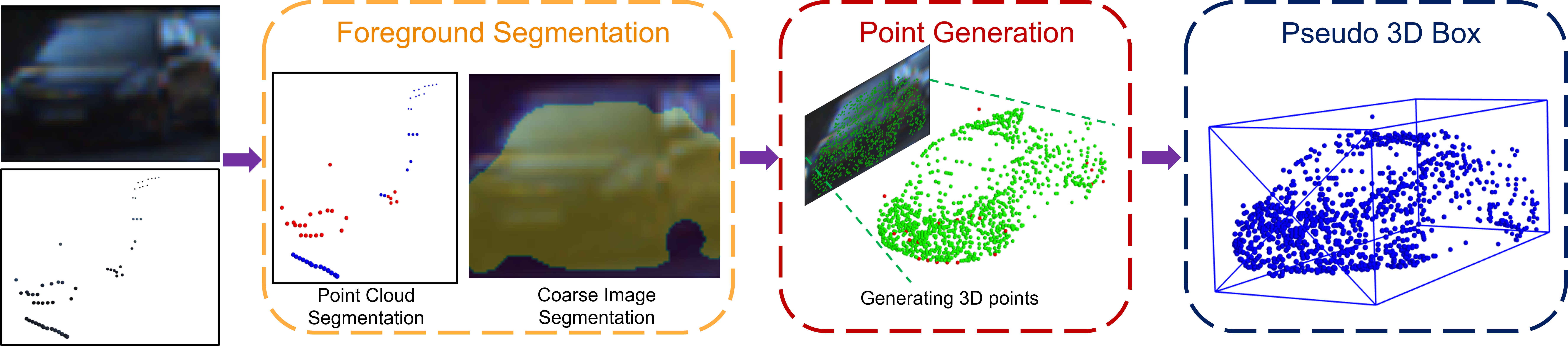}
    \caption{The 3-stages MAP-Gen workflow. First, an image and a point cloud are respectively segmented. Then, the proposed multimodal attention module samples targets from the 2D image and generates new 3D points (in green). In the end, a pseudo 3D bounding box is regressed from the enriched point cloud together with global image features.}
    \vspace{-10pt}
    \label{fig:fig2_workflow}
\end{figure*}

\section{Related Work}
\subsection{Multimodal Approaches for 3D tasks}
LiDAR scans are usually sparse without color information, making it challenging to recognize objects, especially for far regions. While images provide dense color information, the absence of depth information severely restricts 3D tasks like object detection tasks. Therefore, it is a natural idea to combine the two modalities for better performance. 

MV3D~\cite{chen2017mv3d} is a pioneering work to merge LiDAR point clouds with 2D images for 3D object detection. To handle the different viewpoints from the LiDAR's bird's eye view (BEV) and an image, AVOD~\cite{ku2018AVOD} makes use of 3D anchors to bridge the different modalities. Later, Frustum PointNet~\cite{qi2018frustumpnet} generates 2D proposals on the image and then uses cascading PointNet~\cite{qi2017pointnet} to predict 3D boxes. Similarly, FGR~\cite{wei2021fgr} extracts preliminary frustum sub clouds from which 3D annotations are generated. On the other hand, PI-RCNN~\cite{xie2020pircnn} and PointPainting~\cite{vora2020pointpainting} try to directly fuse point cloud features with the image information. PointPainting performs semantic segmentation on 2D images and concatenates the segmentation logits with 3D-point features. Besides, PI-RCNN fuses image features into point clouds with their new PACF module. Other methods, such as EPNet~\cite{huang2020epnet} and ACMNet~\cite{zhao2021ACMNet}, adopt a gating mechanism to control the portion of LiDAR and image information during fusion.

Although these previous works have demonstrated satisfactory results at the time they were published~\cite{yoo20203d-cvf, zhao2021ACMNet}, some later single-modal methods have surpassed them~\cite{shi2020pvrcnn, deng2020voxelrcnn}. Moreover, such extra performance often comes at the expense of higher computation and memory costs due to the extra inputs, thus incurring a higher hurdle for online tasks than their offline counterparts. In fact, how to fully utilize multimodal information remains an open question~\cite{wang2020what_makes}. 

\subsection{Generating 3D bounding boxes from weak labels}
Annotating point clouds is typically time-consuming and laborious~\cite{meng2020weakly, meng2021ws3d_towards}, which significantly hinders the preparation of datasets or the fine-tuning of models on custom scenes. Subsequently, it is of high interest and value to automatically generate 3D labels from weak ones such as 2D bounding boxes. Nevertheless, only a few previous works tried to automatically generate 3D boxes from weak 2D annotations. Ref.~\cite{mccraith2021lifting2d} predicts 3D bounding boxes from raw point clouds and 2D instance segmentation results. SDF~\cite{zakharov2020sdf} uses predefined CAD models to estimate 3D geometry based on 2D bounding boxes and the sparse point clouds. FGR~\cite{wei2021fgr} locates frustum point clouds from 2D boxes, coarsely segments the clouds to remove backgrounds, and then directly computes 3D bounding boxes. FGR is a non-training algorithm but achieves state-of-the-art performance for easy samples on KITTI. WS3D~\cite{meng2020weakly, meng2021ws3d_towards} and Ref.~\cite{lee2018leveraging} propose a new type of weak annotation named center clicks, achieving state-of-the-art performance on moderate/hard samples on KITTI. In particular, users first click on the object center from the image and then adjust the centroid on BEV. Nonetheless, center clicks require annotators to continually shift between images and BEVs. In contrast, 2D bounding boxes are more available, either manually annotated or from various 2D detection models~\cite{qi2018frustumpnet}.

Nonetheless, the aforementioned autolabelers mainly employ image information to narrow down the scope without addressing the sparsity problem, resulting in an apparent accuracy drop for moderate and hard samples, which are usually far and hence sparse. This work is based on the premise that alleviating the sparsity problem can effectively boost the quality of generated annotations, as is verified through extensive experiments.

\section{Methodology}
Given an image cropped by a 2D box and the corresponding frustum point cloud, the goal of MAP-Gen is to automatically generate 3D bounding box annotations. As shown in~\cref{fig:fig2_workflow}, MAP-Gen consists of three stages: 1) foreground segmentation, 2) point cloud enrichment, and 3) 3D box regression. The image and point cloud are first segmented to remove the backgrounds. Then new points are generated by the proposed multimodal attention module to alleviate the sparsity issue. Finally, a simple PointNet~\cite{qi2017pointnet} is used to regress 3D boxes from the enriched point clouds. 

Specifically, MAP-Gen is trained with a small amount (500 frames) of ground truth 3D boxes. After training, the MAP-Gen works as an autolabeler and is used to re-label the KITTI dataset. Given weak annotation of 2D boxes, MAP-Gen outputs pseudo 3D labels which can be used for training any object detection network (e.g., PointRCNN). The following sections elaborate on how MAP-Gen generates 3D boxes from weak 2D annotations.

\subsection{Foreground Segmentation}
\label{sec:Method_foreground_segment}
First, the region enclosed by each 2D bounding box is cropped and resized to a fixed size $H\times W$. Given the LiDAR-camera projection matrix, each point in the 3D LiDAR coordinate system can be projected onto the 2D plane of the camera image. As illustrated in~\cref{fig:1a_frustum_projection}, we extract the frustum sub-cloud $\mathcal{P} \in \mathbb{R}^{n\times 3}$ whose 2D projects $\mathcal{P}_{2d} \in \mathbb{R}^{n\times 2}$ are within the target 2D bounding box. $n$ is the number of points in the cloud. Corresponding RGB values for $\mathcal{P}$ are also extracted based on the projection.

Then, the RGB point cloud is segmented by a PointNet~\cite{qi2017pointnet}, while the image is segmented by a PSPNet~\cite{zhao2017pspnet} with a Res18~\cite{he2016resnet} backbone. The outputs are $l_{pn}, l_{psp} \in \mathbb{R}^{n\times2}$ respectively, denoting the foreground/background segmentation results. Here $l_{psp}$ is extracted from the PSPNet's $H\times W$ segmentation map by the aforementioned LiDAR-image mapping. During training, points within the ground truth 3D boxes are regarded as foreground, and Cross-Entropy Loss is used for both $l_{pn}$ and $l_{psp}$. We balance the loss for foregrounds and backgrounds by their populations.
Although there are no ground-truth segmentation masks for the images, the CNN is partially supervised by calculating loss only for pixels with LiDAR point projection. As shown in~\cref{fig:fig2_workflow}, images are coarsely segmented.

\subsection{Point Generation with Multimodal Attention}
\label{sec:point_generation}
After segmentation, we remove the background points, resulting in a new point cloud $\mathcal{P'}\in \mathbb{R}^{n'\times 3}$, where $n'$($\leq n$) is the number of remaining points. Then the proposed multimodal attention module generates new foreground points to enrich $\mathcal{P'}$ to a predefined size $m$ via the following steps:%

\paragraph{2D Target Sampling}
To enrich $\mathcal{P'}$, $k=m-n'$ pixels are randomly sampled from the image foreground, representing \emph{targets}. The sampled target pixels have known 2D coordinates $\mathcal{C}_{2d} \in \mathbb{R}^{k\times2}$ and features $\mathcal{F}_{2d}\in\mathbb{R}^{k\times {c_i}}$ extracted from the feature pyramid in PSPNet. Their 3D coordinates are unknown and to be estimated in later steps.
\begin{figure*}[t]
    \centering
    \includegraphics[width=\textwidth]{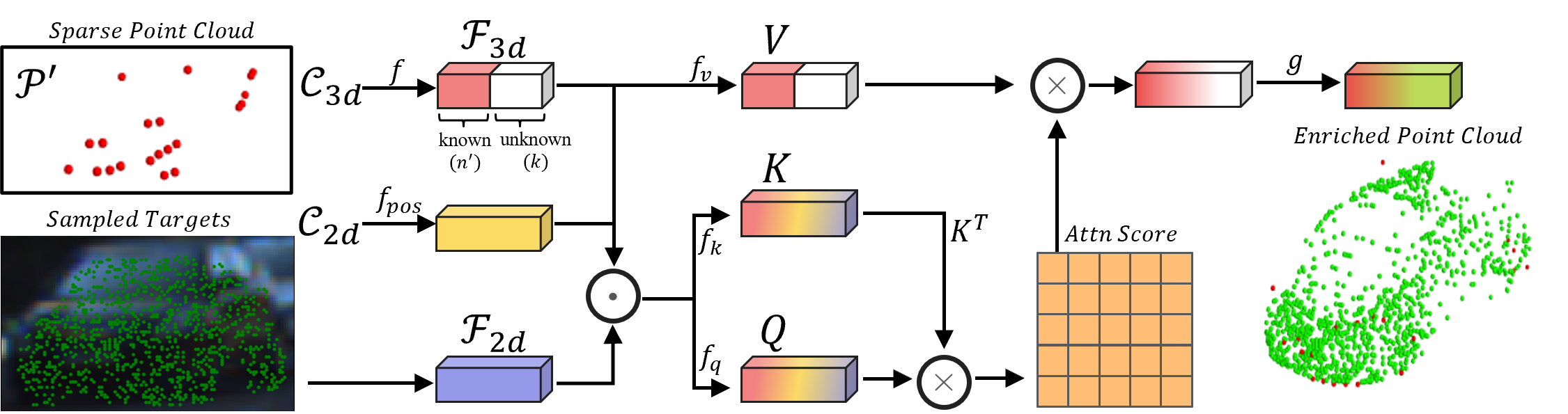}
    \caption{The proposed multimodal attention. Besides the $n'$ foreground points in $\mathcal{P}'$, $k$ extra 2D targets are sampled from the coarse foreground region of the image. All the $n'+k$ points have known 2D features $\mathcal{F}_{2d}$ and coordinates $\mathcal{C}_{2d}$, but only the $n'$ points have known $\mathcal{C}_{3d}$ for calculating $\mathcal{F}_{3d}$. For the $k$ target points, trainable embedding vectors are padded, meaning \emph{unknown} 3D features (marked as white). Via the attention mechanism, targets query their 3D features from the context points based on their semantic and geometric relationships. Finally, the targets' 3D coordinates are decoded, and the point cloud is hence enriched. $\odot$ means concatenation and $\otimes$ means matrix multiplication. Best viewed in color.}
    \label{fig:fig3_multimodal_attention}
\end{figure*}

\paragraph{Sequence Building}
Merging the segmented cloud $\mathcal{P'}$ and the $k$ sampled targets, an order-invariant sequence $\mathcal{S}$ of length $m$ is built,
\begin{gather}
    \mathcal{S}=\left[\left(\mathcal{C}_{2d}^{(1)}, \mathcal{F}_{2d}^{(1)}, \mathcal{F}_{3d}^{(1)}\right), \dots, \left(\mathcal{C}_{2d}^{(m)}, \mathcal{F}_{2d}^{(m)}, \mathcal{F}_{3d}^{(m)}\right)\right] \\ \nonumber
    \mathcal{F}_{3d}^{(i)} = \begin{cases}
    f(C_{3d}^{(i)}),  & i=1,2,...,n'\\
    u, & i=n'+1, ..., m
    \end{cases}
\end{gather}
In $\mathcal{S}$, each tuple denotes a point with 3 properties, namely, 2D coordinate $\mathcal{C}_{2d}^{(i)}$, 2D feature $\mathcal{F}_{2d}^{(i)}$ and 3D feature $\mathcal{F}_{3d}^{(i)}$ derived from 3D coordinates $\mathcal{C}_{3d}^{(i)}$ by the function $f$ (e.g., a multilayer perceptron). For the $k$ targets without known 3D coordinates, we fill their 3D features with an identical trainable embedding vector $u$ as the \emph{unknown} features. Next, we try to recover 3D features and hence 3D coordinates for the $k$ targets.

\paragraph{Multimodal Attention}
\label{section:multi_attn}
We restore the 3D features by the proposed multimodal attention mechanism. By referencing context points based on the semantic and geometric relationships, 3D features are interpolated for the target points. As shown in~\cref{fig:fig3_multimodal_attention}, the multimodal attention module effectively alleviates the sparsity problem by generating new points.

Recently, self-attention approaches have achieved great success in the NLP and 2D CV tasks \cite{devlin2018bert, dosovitskiy2020vit, liu2021swinTransformer, carion2020dert}.
While this work proposes a new type of self-attention to generate 3D points from context points using multimodal inputs. Given a target point's 2D feature and coordinates, we aim to restore its 3D feature $\mathcal{F}_{3d}$ by querying context points from $\mathcal{P}'$. As illustrated in \cref{fig:fig3_multimodal_attention}, for each point we derive a \emph{key} $K$ and a \emph{query} $Q$ from their 2D coordinates, 2D features and 3D (known or unknown) features. The attention score between every two points is calculated by multiplying one point's query with the other's key as in \cref{equation:2d_self-attention}. Each point's representation, viz. \emph{value} $V$, is then updated with the context points which are weighted by the attention scores. The procedure can be formulated as: 
\begin{subequations}
    \begin{align}
        &Q = f_q(f_{pos}(\mathcal{C}_{2d}) \odot \mathcal{F}_{2d} \odot \mathcal{F}_{3d}) \\
        &K = f_k(f_{pos}(\mathcal{C}_{2d}) \odot \mathcal{F}_{2d} \odot \mathcal{F}_{3d}) \\
        &V = f_v(\mathcal{F}_{3d})
        \label{equation:2c_value}\\
        &\mathcal{F}_{3d}^{'} = \sigma(\frac{QK^T}{\sqrt{d}})V%
        \label{equation:2d_self-attention}
    \end{align}%
    \label{equation:multimodal_attention}%
\end{subequations}
where $f_{pos}$ is position embedding function, $\odot$ denotes concatenation, $f_q, f_k, f_v$ are three fully connected layers, $\sigma$ is softmax function and $\sqrt{d}$ is a scaling factor as in Ref.\cite{vaswani2017transformer}.

Noted that key and query are derived from all available information while value comes from the 3D features $\mathcal{F}_{3d}$ only. The motivation of this asymmetric design is that 2D position and semantic information provide critical references to determine the relationship of two points. However, we are only interested in the 3D feature for output to decode the 3D coordinates of target pixels. Therefore, we deliberately avoid mixing different modalities into the point representation $V$.

Different from common transformers, $\mathcal{S}$ is unordered so we do not use sequence-level position embeddings, except the image position embedding $f_{pos}(\mathcal{C}_{2d})$. Moreover, the target points will not be attended by others, because they have no 3D features and do not contribute to the context. After a stack of multimodal attention layers, a multilayer perceptron (MLP) head $g$ is used to retrieve 3D coordinates for the targets:
    \begin{equation} 
        \label{equation:decode_xyz}
        \hat{\mathcal{C}_{3d}} = g(\mathcal{F}_{3d}^{'})
    \end{equation}

\paragraph{Mask and Recover}
There are no point generation labels for our task. Therefore, we adopt a \emph{mask-and-recover} strategy to supervise the multimodal attention module. A random portion of the foreground points from $\mathcal{P}'$ are masked and the model is asked to recover them during training. For the points to be masked, we replace their original 3D features with the trainable embedding vector $u$. The point generation loss $\mathcal{L}_{gen}$ is calculated by SmoothL1 loss on masked points.

\begin{table*}[t]
    \centering
    \begin{tabular}{lccccccc}
\toprule
\multirow{2}{*}{Method} & \multirow{2}{*}{Full Supervision} & \multicolumn{3}{c}{$\text{AP}_{3D}(IoU=0.7)$} & \multicolumn{3}{c}{$\text{AP}_{BEV}(IoU=0.7)$} \\
\cmidrule(lr){3-5} \cmidrule(lr){6-8}
& & \multicolumn{1}{c}{Easy} & \multicolumn{1}{c}{Moderate} & \multicolumn{1}{c}{Hard}
&  \multicolumn{1}{c}{Easy} & \multicolumn{1}{c}{Moderate} & \multicolumn{1}{c}{Hard} \\
\midrule
MV3D\cite{chen2017mv3d} & \cmark & 74.97 & 63.63 & 54.00 & 86.62 & 78.93 & 69.80 \\
F-PointNet\cite{qi2018frustumpnet} & \cmark & 82.19 & 69.79 & 60.59 & 91.17 & 84.67 & 74.77 \\
AVOD\cite{ku2018AVOD} & \cmark & 83.07 & 71.76 & 65.73 & 90.99 & 84.82 & 79.62 \\
SECOND\cite{yan2018second} & \cmark & 83.34 & 72.55 & 65.82 & 89.39 & 83.77 & 78.59 \\
PointPillars\cite{lang2019pointpillars} & \cmark & 82.58 & 74.31 & 68.99 & 90.07 & 86.56 & 82.81 \\
PointRCNN\cite{shi2019pointrcnn} & \cmark & \textit{86.96} & \textit{75.64} & \textit{70.70} & \textit{92.13} & \textit{87.39} & \textit{82.72} \\
Part-A$^2$\cite{shi2020partA2} & \cmark & 87.81 & 78.49 & 73.51 & 91.70 & 87.79 & 84.61 \\
PV-RCNN\cite{shi2020pvrcnn} & \cmark & 90.25 & 81.43 & 76.82 & 94.98 & 90.65 & 86.14 \\
\midrule
FGR\cite{wei2021fgr} & (2D box) & 80.26 & 68.47 & 61.57 & 90.64 & 82.67 & 75.46 \\
WS3D\cite{meng2020weakly} & (BEV Centroid) & 80.15 & 69.64 & 63.71 & 90.11 & 84.02 & 76.97 \\
WS3D(2021)\cite{meng2021ws3d_towards} & (BEV Centroid) & \textit{80.99} & \textit{70.59} & \textit{64.23} & \textit{90.96} & \textit{84.93} & \textit{77.96} \\
\midrule
MAP-Gen (Ours) & (2D box) & \textbf{81.51} & \textbf{74.14} & \textbf{67.55} & \textbf{90.61} & \textbf{85.91} & \textbf{80.58}\\
\bottomrule
    \end{tabular}
    \caption{Results on KITTI test set. We train the proposed MAP-Gen with 500 frames data from the KITTI training set. Then the MAP-Gen generates pseudo labels for all the frames in the training set to train another PointRCNN model.}
    \vspace{-10pt}
    \label{tab:KITTI_test_results}
\end{table*}

\subsection{Regressing 3D boxes}
The proposed multimodal attention module enriches the sparse point cloud $\mathcal{P}'$ and alleviates the sparsity and irregularity issues simultaneously. As in \cref{sec:Method_foreground_segment} the corresponding RGB values are extracted again, forming an enriched RGB cloud $\mathcal{P}_{rgb} \in \mathbb{R}^{m\times 6}$. Another PointNet encodes the $\mathcal{P}_{rgb}$ into a global feature vector $\mathcal{F}_{pc}$. Meanwhile, the global image representation is derived by performing global average pooling $\mathcal{A}$ on the image feature map $Z\in \mathbb{R}^{H \times W \times C_i}$ from the PSPNet in \cref{sec:Method_foreground_segment}. Finally, an MLP head $h$ decodes the 3D bounding box $B$, which can be formulated as:
\begin{subequations}
   \begin{align}
    \mathcal{F}_{img} &= \mathcal{A}(Z) \\
    B &= h(\mathcal{F}_{pc} \odot \mathcal{F}_{img})%
\end{align}%
\end{subequations}%
where $B$ is defined by a 7-dimensional vector including the $x$, $y$, $z$ location, height, length, width and rotation along the $z$-axis. We use the dIoU loss $\mathcal{L}_{iou}$ \cite{iouloss, pytorchiou, zheng2020distanceiou} for training.

\subsection{Training Procedure}
In the above tri-module design, later stage relies on the performance of previous ones, e.g., image segmentation in \cref{sec:Method_foreground_segment} affects the target sampling in \cref{sec:point_generation}. Therefore, we train them in three stages separately. First, the segmentation PointNet and PSPNet are trained with $\mathcal{L}_{seg}$. Then, we train the multimodal attention module with $\mathcal{L}_{gen}$. In the last stage, the final PointNet with box head are trained together by the loss $\mathcal{L}_{iou}$. Each stage is trained for 150 epochs, and modules not under training are frozen.

\section{Experiments}
\label{section:experiments}
The proposed MAP-Gen is trained with only 500 frames of the KITTI dataset, and is then used to relabel the whole dataset given weak annotation of 2D boxes. We examine MAP-Gen in two ways: \begin{enumerate*}
    \item Train another object detection network with pseudo 3D labels generated by MAP-Gen, and measure the detector's accuracy.
    \item Directly compare the generated labels with human annotations.
\end{enumerate*} In the end, we provide visualization results, for qualitative evaluation and interpretability.

\subsection{Experiment Settings}

\paragraph{Dataset} We adopt the KITTI Benchmark Dataset~\cite{Geiger2012CVPR_KITTI} and split the data into training set (3712 frames) and validation set (3769 frames) same as~\cite{shi2019pointrcnn, shi2020pvrcnn, wei2021fgr}. Since MAP-Gen works on objects, we pre-process the dataset and extract the objects inside 2D bounding boxes. Each object is affiliated with a frustum sub-cloud $\mathcal{P}$, the projected 2D coordinates $\mathcal{P}_{2d}$ and the cropped image in the 2D box region. Some objects have no foreground LiDAR points, hence not feasible for point generation or autolabeling. Herein we follow FGR~\cite{wei2021fgr} to filter the dataset and preserve objects with at least 30 points and 5 foreground points for training MAP-Gen.


\paragraph{Implementation Details}
For foreground segmentation and box regression, we use vanilla networks of PointNet~\cite{qi2017pointnet} and PSPNet~\cite{zhao2017pspnet} with a light encoder of Res18. The multimodal attention module contains 4 layers, each having 8 heads, whereas the hidden and intermediate sizes are set to be 512 and 1024, respectively. The 2D position encoding $f_{pos}$, 3D feature embedding function $f$ and two heads of $g, h$ are all implemented with 2-layer MLPs. Both input and enriched cloud size $n, m$ are set as 1024. 

MAP-Gen is trained with a small portion (500 frames) of the KITTI training set. Adam optimizer~\cite{kingma2014adam} is adopted with a learning rate of $5\mathrm{e}{-4}$ and batch size of 32. CosineAnnealing~\cite{loshchilov2016sgdr} scheduler is employed with a 5\% linear warmup. Following common 3D object detection practice and FGR~\cite{wei2021fgr}, Average Precision (AP) and mean Intersection over Union (mIoU) are used as our quality metrics. 

\subsection{Weakly Supervised 3D Detection}
\label{sec:exp_weakly_3d_detection}
We examine MAP-Gen by training another 3D object detection network with the generated annotations. Same as FGR~\cite{wei2021fgr} and WS3D~\cite{meng2020weakly}, we choose PointRCNN~\cite{shi2019pointrcnn} as the object detector. The PointRCNN is trained with pseudo 3D labels generated by MAP-Gen from the weak 2D boxes, and then evaluated with ground-truth human annotations. We compare our method with other baselines on both the KITTI test and validation sets. 

\paragraph{Test Set}
\cref{tab:KITTI_test_results} shows the results of our MAP-Gen and other baselines on the KITTI test set. To the best of our knowledge, only a few existing works study the problem of generating strong labels from weak ones, among which FGR~\cite{wei2021fgr} and WS3D~\cite{meng2020weakly, meng2021ws3d_towards} are the current state-of-the-art methods. FGR requires no training but suffers more accuracy drops than WS3D in moderate and hard samples. However, BEV centroids are used as weak annotations in WS3D, which provides more 3D information than 2D boxes.

As shown in~\cref{tab:KITTI_test_results}, our method outperforms WS3D and FGR significantly. Especially for difficult samples, MAP-Gen improves the 3D detection accuracy $\text{AP}_{3D}$ by 3.55\% and 3.32\% on moderate and hard samples, respectively. These results support our motivation well since moderate and hard samples are usually farther and with sparser clouds than easy ones. Namely, they suffer more from the sparsity issue. With MAP-Gen the sparsity is mitigated by newly generated points, and thus the accuracy improves significantly compared with the baselines. Specifically, the weakly supervised PointRCNN achieves 93.7\%, 98.0\%, and 95.5\% performance of the original model on easy, moderate, and hard categories, respectively.
\begin{table}[t]
    \centering
    \begin{tabular}{lcccc}
\toprule
\multirow{2}{*}{Method} & \multirow{2}{*}{Full Supervision} & \multicolumn{3}{c}{$\text{AP}_{3D}(IoU=0.7)$} \\
\cmidrule(lr){3-5}
& & \multicolumn{1}{c}{Easy} & \multicolumn{1}{c}{Moderate} & \multicolumn{1}{c}{Hard} 
\\
\midrule
PointRCNN\cite{shi2019pointrcnn} & \cmark & 88.88 & 78.63 & 77.38\\
\midrule
WS3D\cite{meng2020weakly} & (BEV Cent.) & 84.04 & 75.10 & 73.29 \\
WS3D(2021)\cite{meng2021ws3d_towards} & (BEV Cent.) & 85.04 & \textit{75.94} & \textit{74.38}\\
FGR\cite{wei2021fgr} & (2D box) & \textit{86.68} & 73.55 & 67.91\\
\midrule
MAP-Gen (Ours) & (2D box) & \textbf{87.87} & \textbf{77.98} & \textbf{76.18}\\
\bottomrule
    \end{tabular}
    \caption{KITTI val set results versus the fully supervised PointRCNN and other weakly supervised baselines.}
    \label{tab:compare_weak_methods_KITTIval}
\end{table}

\paragraph{Validation Set}
In \cref{tab:compare_weak_methods_KITTIval}, MAP-Gen achieves over 98\% of the original accuracy on all difficulty levels. Impressively, the accuracy drop on the validation set is merely 1.2\% for the hard category. On the validation set, MAP-Gen yields comparable quality as human annotations. Compared with previous methods of FGR and WS3D, the improvements are still promising, being 1.19\%, 2.04\% and 1.8\% higher than the second-best, for easy, moderate and hard samples. Noted that another three works CC~\cite{tang2019cc_semi}, VS3D~\cite{qin2020vs3d} and SDF~\cite{zakharov2020sdf} also report their scores on the KITTI validation set. However, they adopt much easier metrics of $\text{AP}_{3D}$(IoU=0.25) or $\text{AP}_{3D}$(IoU=0.5), and have been surpassed by FGR and WS3D by large margins, so their results are not listed in~\cref{tab:compare_weak_methods_KITTIval}. 

The above results demonstrate that MAP-Gen can automatically generate high-quality pseudo 3D annotations from 2D bounding boxes, saving considerable workload for human annotators. Our method also significantly improves the accuracy for weakly supervised 3D detection.
\begin{table}[t]
    \centering
    \begin{tabular}{lcccc}
\toprule
\multicolumn{2}{c}{Variation} & \multirow{2}{*}{mErr} & \multirow{2}{*}{mIoU} & \multirow{2}{*}{Recall 0.7}\\
\cmidrule(lr){1-2}
Item & Variants & & \\
\midrule
MAP-Gen (ours) & - & \textbf{0.1325} & \textbf{67.64}  & \textbf{62.70} \\
\midrule
Generation & \xmark  & - & 65.72 & 57.89 \\
\midrule
\multirow{2}{*}{Pos-Emb} & Sinusoid & 0.1374 & 67.43 & 60.82 \\
& None & 0.1372 & 67.05 & 60.52 \\
\midrule
\multirow{2}{*}{Fusion} & Add & 0.1417 & 67.40 & 62.53 \\
& Gating & 0.1665 & 67.39 & 60.81 \\
\midrule
\multirow{1}{*}{Asymmetric} & \xmark & 0.1537 & 67.29 & 61.42 \\
\bottomrule

    \end{tabular}
    \caption{Ablation study on MAP-Gen. Enriched point cloud effectively improves pseudo labels' quality.}
    \vspace{-5pt}
    \label{tab:ablation study}
\end{table}

\subsection{Ablation study}
We also conduct ablation studies on the KITTI validation set to further verify MAP-Gen's contributions. To decouple possible perturbations from the detector, we compare the pseudo labels directly with the manual counterparts. Since the major innovation of MAP-Gen is point generation with the proposed multimodal attention design, we focus on ablating this module. In~\cref{tab:ablation study}, we evaluate different technical choices. The accuracy is measured in three metrics, mean error of estimated 3D coordinates, mean IoU and recall with $\text{IoU}>0.7$ as the threshold. In the second row of \cref{tab:ablation study}, the multimodal attention module is removed, resulting in a large drop of 4.81\% for recall. Below that, we examine other position embedding functions. \emph{Sinusoid} is the original position embedding function from \cite{vaswani2017transformer} and \emph{None} means no position embedding is used. Different fusion methods, namely \emph{add} and \emph{gating}, are also evaluated. In the last row, we use vanilla self-attention, where $Q, K, V$ have the same information source. The above variants' all have slight accuracy drops, but still obviously higher than without the generation module, proving the effectiveness of our method.

\begin{figure}[t]
    \centering
    \begin{subfigure}{\columnwidth}
        \begin{subfigure}{0.32\columnwidth}
             \centering
             \includegraphics[width=\textwidth]{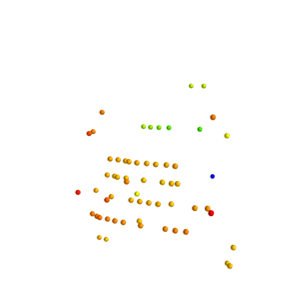}
         \end{subfigure}
         \vspace{1pt}
        \begin{subfigure}{0.32\columnwidth}
             \centering
             \includegraphics[width=\textwidth]{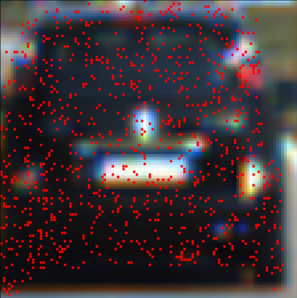}
         \end{subfigure}
         \vspace{1pt}
        \begin{subfigure}{0.32\columnwidth}
             \centering
             \includegraphics[width=\textwidth]{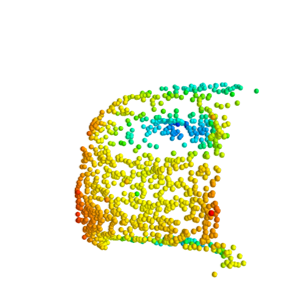}
         \end{subfigure}
         \vspace{1pt}
    \end{subfigure}
     \begin{subfigure}{\columnwidth}
          \begin{subfigure}{0.32\columnwidth}
             \centering
             \includegraphics[width=\textwidth]{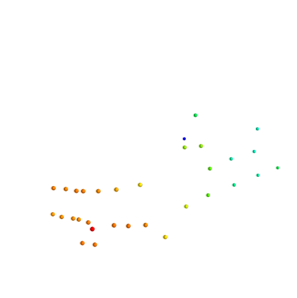}
         \end{subfigure}
         \vspace{1pt}
        \begin{subfigure}{0.32\columnwidth}
             \centering
             \includegraphics[width=\textwidth]{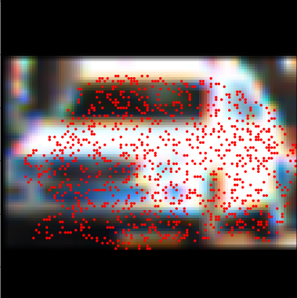}
         \end{subfigure}
         \vspace{1pt}
        \begin{subfigure}{0.32\columnwidth}
             \centering
             \includegraphics[width=\textwidth]{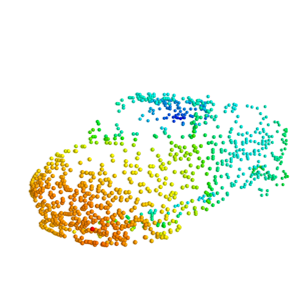}
         \end{subfigure}
         \vspace{1pt}
     \end{subfigure}
     \begin{subfigure}{\columnwidth}
          \begin{subfigure}{0.32\columnwidth}
             \centering
             \includegraphics[width=\textwidth]{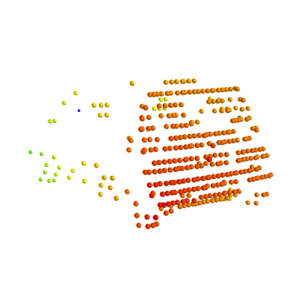}
         \end{subfigure}
        \begin{subfigure}{0.32\columnwidth}
             \centering
             \includegraphics[width=\textwidth]{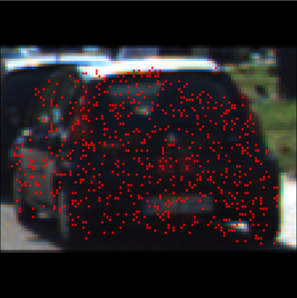}
         \end{subfigure}
        \begin{subfigure}{0.32\columnwidth}
             \centering
             \includegraphics[width=\textwidth]{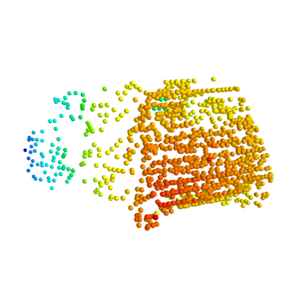}
         \end{subfigure}
     \end{subfigure}
     
    \caption{Visualization of the original point cloud (left), sampled targets (middle) and enriched point cloud (right).}
    \vspace{-10pt}
    \label{fig:visualization}
\end{figure}

\section{Qualitative Results}
In~\cref{fig:visualization}, we visualize some samples including the segmented point clouds $\mathcal{P'}$ (left), images with target pixels (middle), and enriched points cloud from MAP-Gen (right). It can be observed that the problems of sparsity and irregularity have been significantly overcome by MAP-Gen, whereby the improvement is more obvious for sparser samples. This also explains why sparser (moderate/hard) samples benefit more from MAP-Gen than denser (easy) ones.


\section{Conclusion}
This work has proposed MAP-Gen, an autolabeler for generating high-quality 3D annotations from the weak labels of 2D bounding boxes. To overcome the problems of point cloud sparsity and irregularity, a novel multimodal attention mechanism is developed for estimating 3D geometry and enriching the sparse point clouds based on 2D images. Evaluated on the KITTI dataset, MAP-Gen is proved to significantly improve the accuracy of weakly supervised object detectors, surpassing state-of-the-arts by a large margin. MAP-Gen represents a new strategy to enrich sparse point clouds, and has an important implication on a fully automated annotation flow utilizing multimodal information.

\newpage






\bibliographystyle{IEEEtran}
%


\bibliography{ref}

\begin{thebibliography}{10}
\providecommand{\url}[1]{#1}
\csname url@samestyle\endcsname
\providecommand{\newblock}{\relax}
\providecommand{\bibinfo}[2]{#2}
\providecommand{\BIBentrySTDinterwordspacing}{\spaceskip=0pt\relax}
\providecommand{\BIBentryALTinterwordstretchfactor}{4}
\providecommand{\BIBentryALTinterwordspacing}{\spaceskip=\fontdimen2\font plus
\BIBentryALTinterwordstretchfactor\fontdimen3\font minus
  \fontdimen4\font\relax}
\providecommand{\BIBforeignlanguage}[2]{{%
\expandafter\ifx\csname l@#1\endcsname\relax
\typeout{** WARNING: IEEEtran.bst: No hyphenation pattern has been}%
\typeout{** loaded for the language `#1'. Using the pattern for}%
\typeout{** the default language instead.}%
\else
\language=\csname l@#1\endcsname
\fi
#2}}
\providecommand{\BIBdecl}{\relax}
\BIBdecl

\bibitem{Geiger2012CVPR_KITTI}
A.~Geiger, P.~Lenz, and R.~Urtasun, ``Are we ready for autonomous driving? the
  kitti vision benchmark suite,'' in \emph{Conference on Computer Vision and
  Pattern Recognition (CVPR)}, 2012.

\bibitem{sun2020scalability_waymo}
P.~Sun, H.~Kretzschmar, X.~Dotiwalla, A.~Chouard, V.~Patnaik, P.~Tsui, J.~Guo,
  Y.~Zhou, Y.~Chai, B.~Caine \emph{et~al.}, ``Scalability in perception for
  autonomous driving: Waymo open dataset,'' in \emph{Proceedings of the
  IEEE/CVF Conference on Computer Vision and Pattern Recognition}, 2020, pp.
  2446--2454.

\bibitem{caesar2020nuscenes}
H.~Caesar, V.~Bankiti, A.~H. Lang, S.~Vora, V.~E. Liong, Q.~Xu, A.~Krishnan,
  Y.~Pan, G.~Baldan, and O.~Beijbom, ``nuscenes: A multimodal dataset for
  autonomous driving,'' in \emph{Proceedings of the IEEE/CVF Conference on
  Computer Vision and Pattern Recognition}, 2020, pp. 11\,621--11\,631.

\bibitem{pandaset}
P.~Xiao, Z.~Shao, S.~Hao, Z.~Zhang, X.~Chai, J.~Jiao, Z.~Li, J.~Wu, K.~Sun,
  K.~Jiang, Y.~Wang, and D.~Yang, ``Pandaset: Advanced sensor suite dataset for
  autonomous driving,'' in \emph{2021 IEEE International Intelligent
  Transportation Systems Conference (ITSC)}, 2021, pp. 3095--3101.

\bibitem{huang2019apolloscape}
X.~Huang, P.~Wang, X.~Cheng, D.~Zhou, Q.~Geng, and R.~Yang, ``The apolloscape
  open dataset for autonomous driving and its application,'' \emph{IEEE
  Transactions on Pattern Analysis and Machine Intelligence}, vol.~42, no.~10,
  pp. 2702--2719, 2019.

\bibitem{song2015sun_rgbd}
S.~Song, S.~P. Lichtenberg, and J.~Xiao, ``Sun rgb-d: A rgb-d scene
  understanding benchmark suite,'' in \emph{Proceedings of the IEEE Conference
  on Computer Vision and Pattern Recognition}, 2015, pp. 567--576.

\bibitem{meng2020weakly}
Q.~Meng, W.~Wang, T.~Zhou, J.~Shen, L.~Van~Gool, and D.~Dai, ``Weakly
  supervised 3d object detection from lidar point cloud,'' in \emph{European
  Conference on Computer Vision}.\hskip 1em plus 0.5em minus 0.4em\relax
  Springer, 2020, pp. 515--531.

\bibitem{bearman2016s_centerclick}
A.~Bearman, O.~Russakovsky, V.~Ferrari, and L.~Fei-Fei, ``What’s the point:
  Semantic segmentation with point supervision,'' in \emph{European Conference
  on Computer Vision}.\hskip 1em plus 0.5em minus 0.4em\relax Springer, 2016,
  pp. 549--565.

\bibitem{maninis2018deep_extremecut}
K.-K. Maninis, S.~Caelles, J.~Pont-Tuset, and L.~Van~Gool, ``Deep extreme cut:
  From extreme points to object segmentation,'' in \emph{Proceedings of the
  IEEE Conference on Computer Vision and Pattern Recognition}, 2018, pp.
  616--625.

\bibitem{papadopoulos2017extreme_click}
D.~P. Papadopoulos, J.~R. Uijlings, F.~Keller, and V.~Ferrari, ``Extreme
  clicking for efficient object annotation,'' in \emph{Proceedings of the IEEE
  International Conference on Computer Vision}, 2017, pp. 4930--4939.

\bibitem{castrejon2017polygon_rnn}
L.~Castrejon, K.~Kundu, R.~Urtasun, and S.~Fidler, ``Annotating object
  instances with a polygon-rnn,'' in \emph{Proceedings of the IEEE Conference
  on Computer Vision and Pattern Recognition}, 2017, pp. 5230--5238.

\bibitem{acuna2018polygon_rnnpp}
D.~Acuna, H.~Ling, A.~Kar, and S.~Fidler, ``Efficient interactive annotation of
  segmentation datasets with polygon-rnn++,'' in \emph{Proceedings of the IEEE
  Conference on Computer Vision and Pattern Recognition}, 2018, pp. 859--868.

\bibitem{wei2021fgr}
Y.~Wei, S.~Su, J.~Lu, and J.~Zhou, ``Fgr: Frustum-aware geometric reasoning for
  weakly supervised 3d vehicle detection,'' \emph{arXiv preprint
  arXiv:2105.07647}, 2021.

\bibitem{qi2018frustumpnet}
C.~R. Qi, W.~Liu, C.~Wu, H.~Su, and L.~J. Guibas, ``Frustum pointnets for 3d
  object detection from rgb-d data,'' in \emph{Proceedings of the IEEE
  Conference on Computer Vision and Pattern Recognition}, 2018, pp. 918--927.

\bibitem{mccraith2021lifting2d}
R.~McCraith, E.~Insafutdinov, L.~Neumann, and A.~Vedaldi, ``Lifting 2d object
  locations to 3d by discounting lidar outliers across objects and views,''
  \emph{arXiv preprint arXiv:2109.07945}, 2021.

\bibitem{wilson20203d_for_free_hdmap}
B.~Wilson, Z.~Kira, and J.~Hays, ``3d for free: Crossmodal transfer learning
  using hd maps,'' \emph{arXiv preprint arXiv:2008.10592}, 2020.

\bibitem{vora2020pointpainting}
S.~Vora, A.~H. Lang, B.~Helou, and O.~Beijbom, ``Pointpainting: Sequential
  fusion for 3d object detection,'' in \emph{Proceedings of the IEEE/CVF
  Conference on Computer Vision and Pattern Recognition}, 2020, pp. 4604--4612.

\bibitem{meng2021ws3d_towards}
Q.~Meng, W.~Wang, T.~Zhou, J.~Shen, Y.~Jia, and L.~Van~Gool, ``Towards a weakly
  supervised framework for 3d point cloud object detection and annotation,''
  \emph{IEEE Transactions on Pattern Analysis and Machine Intelligence}, 2021.

\bibitem{qin2020vs3d}
Z.~Qin, J.~Wang, and Y.~Lu, ``Weakly supervised 3d object detection from point
  clouds,'' in \emph{Proceedings of the 28th ACM International Conference on
  Multimedia}, 2020, pp. 4144--4152.

\bibitem{lang2019pointpillars}
A.~H. Lang, S.~Vora, H.~Caesar, L.~Zhou, J.~Yang, and O.~Beijbom,
  ``Pointpillars: Fast encoders for object detection from point clouds,'' in
  \emph{Proceedings of the IEEE/CVF Conference on Computer Vision and Pattern
  Recognition}, 2019, pp. 12\,697--12\,705.

\bibitem{zhou2018voxelnet}
Y.~Zhou and O.~Tuzel, ``Voxelnet: End-to-end learning for point cloud based 3d
  object detection,'' in \emph{Proceedings of the IEEE Conference on Computer
  Vision and Pattern Recognition}, 2018, pp. 4490--4499.

\bibitem{qi2017pointnet}
C.~R. Qi, H.~Su, K.~Mo, and L.~J. Guibas, ``Pointnet: Deep learning on point
  sets for 3d classification and segmentation,'' in \emph{Proceedings of the
  IEEE Conference on Computer Vision and Pattern Recognition}, 2017, pp.
  652--660.

\bibitem{qi2017pointnet++}
C.~R. Qi, L.~Yi, H.~Su, and L.~J. Guibas, ``Pointnet++: Deep hierarchical
  feature learning on point sets in a metric space,'' \emph{arXiv preprint
  arXiv:1706.02413}, 2017.

\bibitem{chen2017mv3d}
X.~Chen, H.~Ma, J.~Wan, B.~Li, and T.~Xia, ``Multi-view 3d object detection
  network for autonomous driving,'' in \emph{Proceedings of the IEEE Conference
  on Computer Vision and Pattern Recognition}, 2017, pp. 1907--1915.

\bibitem{ku2018AVOD}
J.~Ku, M.~Mozifian, J.~Lee, A.~Harakeh, and S.~L. Waslander, ``Joint 3d
  proposal generation and object detection from view aggregation,'' in
  \emph{2018 IEEE/RSJ International Conference on Intelligent Robots and
  Systems (IROS)}.\hskip 1em plus 0.5em minus 0.4em\relax IEEE, 2018, pp. 1--8.

\bibitem{xie2020pircnn}
L.~Xie, C.~Xiang, Z.~Yu, G.~Xu, Z.~Yang, D.~Cai, and X.~He, ``Pi-rcnn: An
  efficient multi-sensor 3d object detector with point-based attentive
  cont-conv fusion module,'' in \emph{Proceedings of the AAAI Conference on
  Artificial Intelligence}, vol.~34, no.~07, 2020, pp. 12\,460--12\,467.

\bibitem{huang2020epnet}
T.~Huang, Z.~Liu, X.~Chen, and X.~Bai, ``Epnet: Enhancing point features with
  image semantics for 3d object detection,'' in \emph{European Conference on
  Computer Vision}.\hskip 1em plus 0.5em minus 0.4em\relax Springer, 2020, pp.
  35--52.

\bibitem{zhao2021ACMNet}
S.~Zhao, M.~Gong, H.~Fu, and D.~Tao, ``Adaptive context-aware multi-modal
  network for depth completion,'' \emph{IEEE Transactions on Image Processing},
  2021.

\bibitem{yoo20203d-cvf}
J.~H. Yoo, Y.~Kim, J.~Kim, and J.~W. Choi, ``3d-cvf: Generating joint camera
  and lidar features using cross-view spatial feature fusion for 3d object
  detection,'' in \emph{Computer Vision--ECCV 2020: 16th European Conference,
  Glasgow, UK, August 23--28, 2020, Proceedings, Part XXVII 16}.\hskip 1em plus
  0.5em minus 0.4em\relax Springer, 2020, pp. 720--736.

\bibitem{shi2020pvrcnn}
S.~Shi, C.~Guo, L.~Jiang, Z.~Wang, J.~Shi, X.~Wang, and H.~Li, ``Pv-rcnn:
  Point-voxel feature set abstraction for 3d object detection,'' in
  \emph{Proceedings of the IEEE/CVF Conference on Computer Vision and Pattern
  Recognition}, 2020, pp. 10\,529--10\,538.

\bibitem{deng2020voxelrcnn}
J.~Deng, S.~Shi, P.~Li, W.~Zhou, Y.~Zhang, and H.~Li, ``Voxel r-cnn: Towards
  high performance voxel-based 3d object detection,'' \emph{arXiv preprint
  arXiv:2012.15712}, 2020.

\bibitem{wang2020what_makes}
W.~Wang, D.~Tran, and M.~Feiszli, ``What makes training multi-modal
  classification networks hard?'' in \emph{Proceedings of the IEEE/CVF
  Conference on Computer Vision and Pattern Recognition}, 2020, pp.
  12\,695--12\,705.

\bibitem{zakharov2020sdf}
S.~Zakharov, W.~Kehl, A.~Bhargava, and A.~Gaidon, ``Autolabeling 3d objects
  with differentiable rendering of sdf shape priors,'' in \emph{Proceedings of
  the IEEE/CVF Conference on Computer Vision and Pattern Recognition}, 2020,
  pp. 12\,224--12\,233.

\bibitem{lee2018leveraging}
J.~Lee, S.~Walsh, A.~Harakeh, and S.~L. Waslander, ``Leveraging pre-trained 3d
  object detection models for fast ground truth generation,'' in \emph{2018
  21st International Conference on Intelligent Transportation Systems
  (ITSC)}.\hskip 1em plus 0.5em minus 0.4em\relax IEEE, 2018, pp. 2504--2510.

\bibitem{zhao2017pspnet}
H.~Zhao, J.~Shi, X.~Qi, X.~Wang, and J.~Jia, ``Pyramid scene parsing network,''
  in \emph{Proceedings of the IEEE Conference on Computer Vision and Pattern
  Recognition}, 2017, pp. 2881--2890.

\bibitem{he2016resnet}
K.~He, X.~Zhang, S.~Ren, and J.~Sun, ``Deep residual learning for image
  recognition,'' in \emph{Proceedings of the IEEE Conference on Computer Vision
  and Pattern Recognition}, 2016, pp. 770--778.

\bibitem{devlin2018bert}
J.~Devlin, M.-W. Chang, K.~Lee, and K.~Toutanova, ``Bert: Pre-training of deep
  bidirectional transformers for language understanding,'' \emph{arXiv preprint
  arXiv:1810.04805}, 2018.

\bibitem{dosovitskiy2020vit}
A.~Dosovitskiy, L.~Beyer, A.~Kolesnikov, D.~Weissenborn, X.~Zhai,
  T.~Unterthiner, M.~Dehghani, M.~Minderer, G.~Heigold, S.~Gelly \emph{et~al.},
  ``An image is worth 16x16 words: Transformers for image recognition at
  scale,'' \emph{arXiv preprint arXiv:2010.11929}, 2020.

\bibitem{liu2021swinTransformer}
Z.~Liu, Y.~Lin, Y.~Cao, H.~Hu, Y.~Wei, Z.~Zhang, S.~Lin, and B.~Guo, ``Swin
  transformer: Hierarchical vision transformer using shifted windows,''
  \emph{arXiv preprint arXiv:2103.14030}, 2021.

\bibitem{carion2020dert}
N.~Carion, F.~Massa, G.~Synnaeve, N.~Usunier, A.~Kirillov, and S.~Zagoruyko,
  ``End-to-end object detection with transformers,'' in \emph{European
  Conference on Computer Vision}.\hskip 1em plus 0.5em minus 0.4em\relax
  Springer, 2020, pp. 213--229.

\bibitem{vaswani2017transformer}
A.~Vaswani, N.~Shazeer, N.~Parmar, J.~Uszkoreit, L.~Jones, A.~N. Gomez,
  {\L}.~Kaiser, and I.~Polosukhin, ``Attention is all you need,'' in
  \emph{Advances in Neural Information Processing Systems}, 2017, pp.
  5998--6008.

\bibitem{yan2018second}
Y.~Yan, Y.~Mao, and B.~Li, ``Second: Sparsely embedded convolutional
  detection,'' \emph{Sensors}, vol.~18, no.~10, p. 3337, 2018.

\bibitem{shi2019pointrcnn}
S.~Shi, X.~Wang, and H.~Li, ``Pointrcnn: 3d object proposal generation and
  detection from point cloud,'' in \emph{Proceedings of the IEEE/CVF Conference
  on Computer Vision and Pattern Recognition}, 2019, pp. 770--779.

\bibitem{shi2020partA2}
S.~Shi, Z.~Wang, J.~Shi, X.~Wang, and H.~Li, ``From points to parts: 3d object
  detection from point cloud with part-aware and part-aggregation network,''
  \emph{IEEE Transactions on Pattern Analysis and Machine Intelligence}, 2020.

\bibitem{iouloss}
D.~{Zhou}, J.~{Fang}, X.~{Song}, C.~{Guan}, J.~{Yin}, Y.~{Dai}, and R.~{Yang},
  ``Iou loss for 2d/3d object detection,'' in \emph{2019 International
  Conference on 3D Vision (3DV)}, 2019, pp. 85--94.

\bibitem{pytorchiou}
lilanxiao, ``Differentiable iou of oriented boxes,''
  \url{https://github.com/lilanxiao/Rotated_IoU}, 2021.

\bibitem{zheng2020distanceiou}
Z.~Zheng, P.~Wang, W.~Liu, J.~Li, R.~Ye, and D.~Ren, ``Distance-iou loss:
  Faster and better learning for bounding box regression,'' in
  \emph{Proceedings of the AAAI Conference on Artificial Intelligence},
  vol.~34, no.~07, 2020, pp. 12\,993--13\,000.

\bibitem{kingma2014adam}
D.~P. Kingma and J.~Ba, ``Adam: A method for stochastic optimization,''
  \emph{arXiv preprint arXiv:1412.6980}, 2014.

\bibitem{loshchilov2016sgdr}
I.~Loshchilov and F.~Hutter, ``Sgdr: Stochastic gradient descent with warm
  restarts,'' \emph{arXiv preprint arXiv:1608.03983}, 2016.

\bibitem{tang2019cc_semi}
Y.~S. Tang and G.~H. Lee, ``Transferable semi-supervised 3d object detection
  from rgb-d data,'' in \emph{Proceedings of the IEEE/CVF International
  Conference on Computer Vision}, 2019, pp. 1931--1940.

\end{thebibliography}

\end{document}